\documentclass[12pt]{article} % Default font size is 12pt, it can be changed here

\usepackage{geometry} % Required to change the page size to A4
\usepackage{amsmath}
\usepackage{graphicx} % Required for including pictures

\usepackage{float} % Allows putting an [H] in \begin{figure} to specify the exact location of the figure
\usepackage{wrapfig} % Allows in-line images such as the example fish picture

\usepackage{lipsum} % Used for inserting dummy 'Lorem ipsum' text into the template

\graphicspath{{./Pictures/}} % Specifies the directory where pictures are stored

\begin{document}

%----------------------------------------------------------------------------------------
%	TITLE PAGE
%----------------------------------------------------------------------------------------
\newcommand{\HRule}{\rule{0.9\linewidth}{0.5mm}} % Defines a new command for the horizontal lines, change thickness here

\begin{center}
%\HRule \\[0.4cm]
{ \Large \bfseries Dense~Associative~Memory~is~Robust~to~Adversarial~Inputs} \\[0.10cm] % Title of your document
\vspace{0.3cm}
%\textsc{\large Dmitry Krotov,\ John Hopfield}\\[0.15cm]
\textsc{\large Dmitry Krotov\footnote{Simons Center for Systems Biology, Institute for Advanced Study, Princeton, NJ, 08540, USA, \  krotov@ias.edu},\ John J Hopfield\footnote{Princeton Neuroscience Institute, Princeton University, Princeton, NJ, 08544, USA, \ hopfield@princeton.edu}}\\[0.15cm]
\vspace{0.3cm}
\textsc{\large Abstract}\\[0.15cm]
\vspace{-0.0cm}
\end{center}

Deep neural networks (DNN) trained in a supervised way suffer from two known problems. First, the minima of the objective function used in learning correspond to data points (also known as rubbish examples or fooling images) that lack semantic similarity with the training data. Second, a clean input can be changed by a small, and often imperceptible for human vision, perturbation, so that the resulting deformed input is misclassified by the network. These findings emphasize the differences between the ways DNN and humans classify patterns, and raise a question of designing learning algorithms that more accurately mimic human perception compared to the existing methods.  
  
Our paper examines these questions within the framework of Dense Associative Memory (DAM) models. These models are defined by the energy function, with higher order (higher than quadratic) interactions between the neurons. We show that in the limit when the power of the interaction vertex in the energy function is sufficiently large, these models have the following three properties. First, the minima of the objective function are free from rubbish images, so that each minimum is a semantically meaningful pattern. Second, artificial patterns poised precisely at the decision boundary look ambiguous to human subjects and share aspects of both classes that are separated by that decision boundary. Third, adversarial images constructed by models with small power of the interaction vertex, which are equivalent to DNN with rectified linear units (ReLU), fail to transfer to and fool the models with higher order interactions. This opens up a possibility to use higher order models for detecting and stopping malicious adversarial attacks. The presented results suggest that DAM with higher order energy functions are closer to human visual perception than DNN with ReLUs.   
  
\vspace{0.2cm}
\HRule

\section{Introduction}
Deep neural networks are highly sensitive to small well-designed perturbations of the inputs, known as adversarial perturbations, which lead to misclassifications of these perturbed inputs. Consider images as an example of the data. A raw image, which is correctly classified by the neural network, can be modified in a small (and often imperceptible to our vision) way, so that the resulting deformed image is classified as a different class \cite{Szegedy}. A related problem is that DNN classify certain images as belonging to some class, although they are unrecognizable to humans as exemplars of any class \cite{Nguyen}. These rubbish or fooling images correspond to patches of the image space that have a small value of the objective function used in training and are located far away from any of the training data. These observations has challenged the integrity of DNN classification and has led to an opinion that their predictions are untrustworthy (see discussion of this issue in numerous blogs), though similar problems are shared by many other machine learning techniques, such as logistic regression, support vector machines, K-nearest neighbors and others \cite{transferability}.  

Further investigation have shown that both adversarial images and rubbish images can be transferred between many different models having distinct architectures, different hyperparameters, and even trained on different training sets \cite{Szegedy,Nguyen}. Moreover, some of the adversarial and rubbish images can be transferred between a diverse set of models in machine learning \cite{transferability}. This opens up a possibility for a potential adversarial attack, when a hacker can train his own model and create a set of images that are misclassified by it, and then deploy this set of images against another victim model, which will also misclassify them. Realistic attacks were studied in \cite{transferability,practical_attack,Kurakin}. Often they do not require any internal knowledge of the victim model. 

In addition to being a security issue, transferability suggests that various computational models learn very similar representations of the data. It also suggests that in order to address these problems, one might have to design training algorithms that learn a very different representation of the data compared to the existing methods. An ideal solution to these issues should be an algorithm that assigns small values of the objective function only to those areas of the image space that are recognizable by humans as images of the corresponding class. It should also require a substantial and recognizable by humans deformation of the initial correctly classified image towards a different target class before the label changes. In spite of a substantial amount of work on these problems \cite{Miyato,non-linearity,strong adversary,Nokland,Random Feature Null}, no algorithm have been identified so far which satisfies these requirements and at the same time is competitive to state of the art algorithms in terms of classification accuracy. 

In a recent paper \cite{Krotov Hopfield} it was proposed that Dense Associative Memory (DAM) models with higher order interactions in the energy function learn representations of the data, which strongly depend on the power of the interaction vertex. The network extracts features form the data for small values of this power, but as the power of the interaction vertex is increased there is a gradual shift to a prototype-based representation, the two extreme regimes of pattern recognition known in cognitive psychology. Remarkably, there is a broad range of powers of the energy function, for which the representation of the data is already in the prototype regime, but the accuracy of classification is still competitive to the best available algorithms (based on DNN with ReLUs). This suggests that the DAM models might behave very differently compared to the standard methods used in deep learning with respect to adversarial deformations. 

In the present paper we report three main results. First, using a gradient decent in the pixel space, a set of ``rubbish'' images is constructed that correspond to the minima of the objective function used in training. This is done on the MNIST dataset of handwritten digits using different values of the power of the interaction vertex, which is denoted by $n$. For small values of the power $n$ these images indeed look like speckled rubbish noise, and do not have any semantic content for human vision, a result consistent with \cite{Nguyen}. However, as the power of the interaction vertex is increased the images gradually become less speckled and more semantically meaningful. In the limit of very large $n\approx20...30$, these images are no longer rubbish at all. They represent plausible images of handwritten digits that could have possibly been produced by a human. Second, starting from clean images from the dataset a set of adversarial images is constructed in such a way that each image is placed exactly on the decision boundary between two label classes. For small powers $n$ these images look very similar to the initial clean image with a little bit of speckled noise added, but are misclassified by the neural network, a result consistent with \cite{Szegedy}. However, as the power of the interaction vertex is increased these adversarial images become less and less similar to the initial clean image. In the limit of very large powers these adversarial images look either like a morphed image of two digits (the initial clean image and another digit from the class that the deformation targets), or the initial digit superimposed on a ``ghost'' image from the target class. Either way the interpretation of the artificial patterns generated by the neural net on the decision boundary requires the presence of another digit from the target class in addition to the initial seed from the dataset, and cannot be explained by simply adding noise to the initial clean image. Third, adversarial and rubbish images generated by models with small $n$ can be transferred to and fool another model with small (but possibly different) value $n$. However, they fail to transfer to models with large $n$. Thus rubbish and adversarial images generated by models with small $n$ cannot fool models with large $n$. In contrast, the ``rubbish'' images generated by models with large $n$ can be transferred to models with small $n$, but this is not a problem since those ``rubbish'' images are actually not rubbish at all and look like credible handwritten digits. These results suggest that the DAMs with a large power of the interaction vertex in the energy function better mimic the psychology of human visual perception than DAMs with a small power. The latter are equivalent to DNNs with ReLUs \cite{Krotov Hopfield}.

 \section{Data representation in $\text{DAM}_n$ networks} \label{DAM networks}
In order to illustrate these results in the simplest possible setting we trained a family of DAM networks for several values of the power of the energy function on the MNIST pixel permutation invariant task. 
\begin{figure}[h]
\begin{center}
\includegraphics[width = 0.3\linewidth]{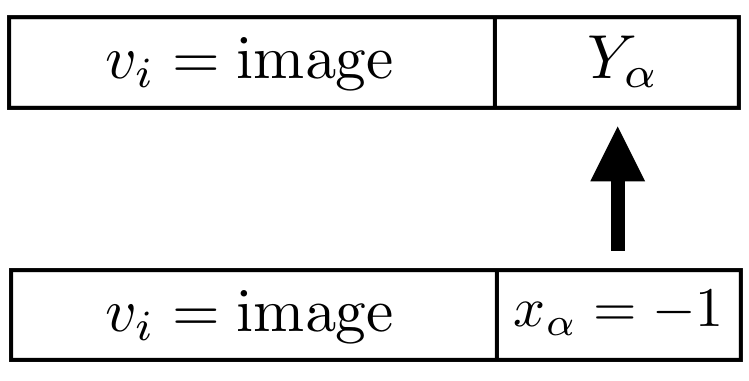}
\end{center}
\caption{\footnotesize{Architecture of the neural network. Visible neurons $v_i$ are equal to the intensities of the pixels, so that the $8$-bit intensities are linearly mapped onto the segment $[-1, +1]$. There are $N=784$ visible neurons $v_i$, and $N_c=10$ classification neurons, which are initialized in the ``off'' state $x_\alpha=-1$ and then updated once to the final state $Y_\alpha$ using (\ref{update_rule}). The model has $K=2000$ memories, with visible parts $\xi^\mu_i$ and recognition part $\xi^\mu_\alpha$, index $\mu=1...K$. Parameter $\beta$ regulates the slope of the $\tanh$ function.}}\label{architecture}
\vspace{-0.0cm}
\end{figure}
 The model is defined by  a set of weights $\xi^\mu_i$ and $\xi^\mu_\alpha$ and the feedforward update rule
 \begin{equation}
Y_\alpha=\tanh\Big[ \beta \Big(\sum\limits_{\mu=1}^K F_n\big(\xi^\mu_i v_i + \xi^\mu_\alpha - \sum\limits_{\gamma \neq \alpha} \xi^\mu_\gamma  \big) -F_n\big(\xi^\mu_i v_i - \xi^\mu_\alpha - \sum\limits_{\gamma \neq \alpha} \xi^\mu_\gamma\big)  \Big)\Big] \label{update_rule}
\end{equation}
where summation over repeated index $i$ is assumed. The functions $F_n$ are rectified polynomials~\cite{Krotov Hopfield}
\begin{equation}\label{rect_polynom} 
 F_n(x)=\begin{cases}x^n,  \ \ x\geq0\\
0, \ \ \ x<0
\end{cases}
\end{equation} 
As we explained in \cite{Krotov Hopfield}, the argument of the update rule (\ref{update_rule}) is the difference of two energies, corresponding to the initial and the final states of the neurons. For this reason the functions $F_n(x)$ are called the energy functions in the rest of the paper, and the integer $n$ is called the power of the interaction vertex. The weights are learned using a backpropagation algorithm (see Appendix A of \cite{Krotov Hopfield}) starting from random initial seeds by minimizing the objective function
\begin{equation}
C=\sum\limits_{\begin{array}{c} \text{\footnotesize{training}}\\ \text{\footnotesize{examples}} \end{array}} \sum\limits_{\alpha=1}^{N_c} \big(Y_\alpha - t_\alpha\big)^{2m}, \label{objective_function}
\end{equation}
where $t_\alpha$ is the target output ($t_\alpha=-1$ for the wrong classes and $t_\alpha=+1$ for the correct class). Hyperparameter $m$ regulates the shape of the objective function.  
    
\begin{figure}[h]
\begin{center}
\includegraphics[width = 0.99\linewidth]{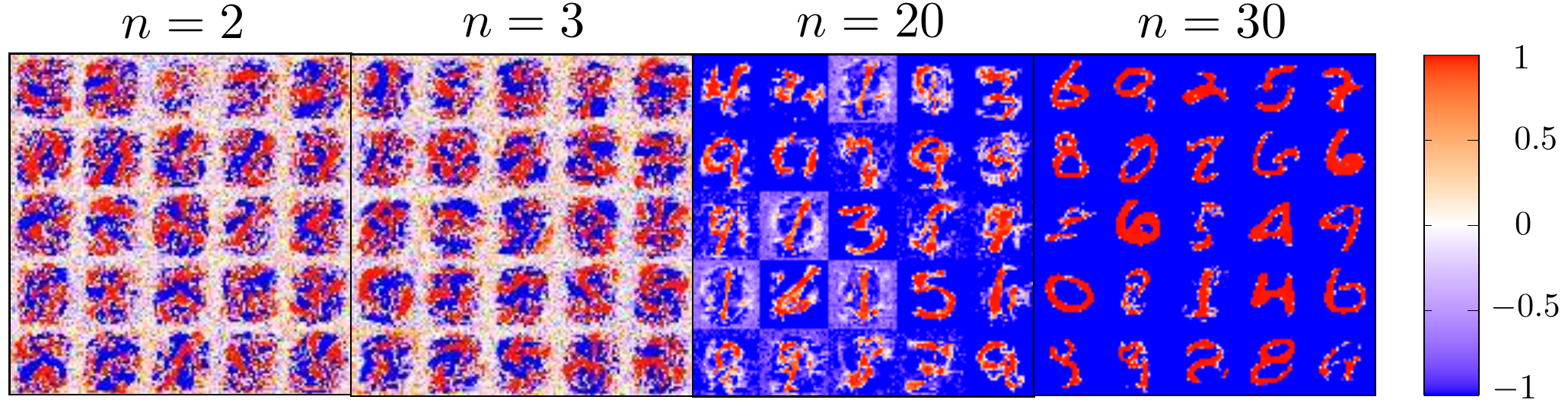}
\end{center}
\caption{\footnotesize{Feature to prototype transition for powers of the energy function $n=2, 3, 20, 30$. For each model 25 randomly selected feature (or prototype) detectors $\xi^\mu_i$ are shown. The value of the $i$-th element of a detector is plotted in the location of the $i$-th pixel to which it couples in the update rule (\ref{update_rule}). The weights are normalized so that $\-1\leq \xi^\mu_i \leq 1$. The color code is explained by the color bar. }}\label{feature_to_prototype}
\end{figure}

 In \cite{Krotov Hopfield} we investigated this model for varying powers of the energy functions $F_n$ and discovered that the network learns feature-based representations of the data for small $n$ and prototype-based representations for large $n$. This is clear from the Fig.\ref{feature_to_prototype}. For small $n$ each feature detector $\xi^\mu_i$ describes a feature which is useful for recognizing several different digit classes. 
As the power $n$ is increased, most of the feature detectors specialize and become responsible for recognizing one possible prototype, which is a constructed representation and is not simply a copy of one image from the training set. Importantly, throughout the range $n=2...30$ the classification accuracy remains approximately the same, $\text{error}_{\text{test}}=1.4 - 1.8\%$. This is comparable to the best result $1.6\%$, achieved by the standard DNNs trained with backpropagation alone \cite{Platt}.

\section{Rubbish examples in $\text{DAM}_n$ networks}  
\begin{figure}[h]
\begin{center}
\includegraphics[width = 0.7\linewidth]{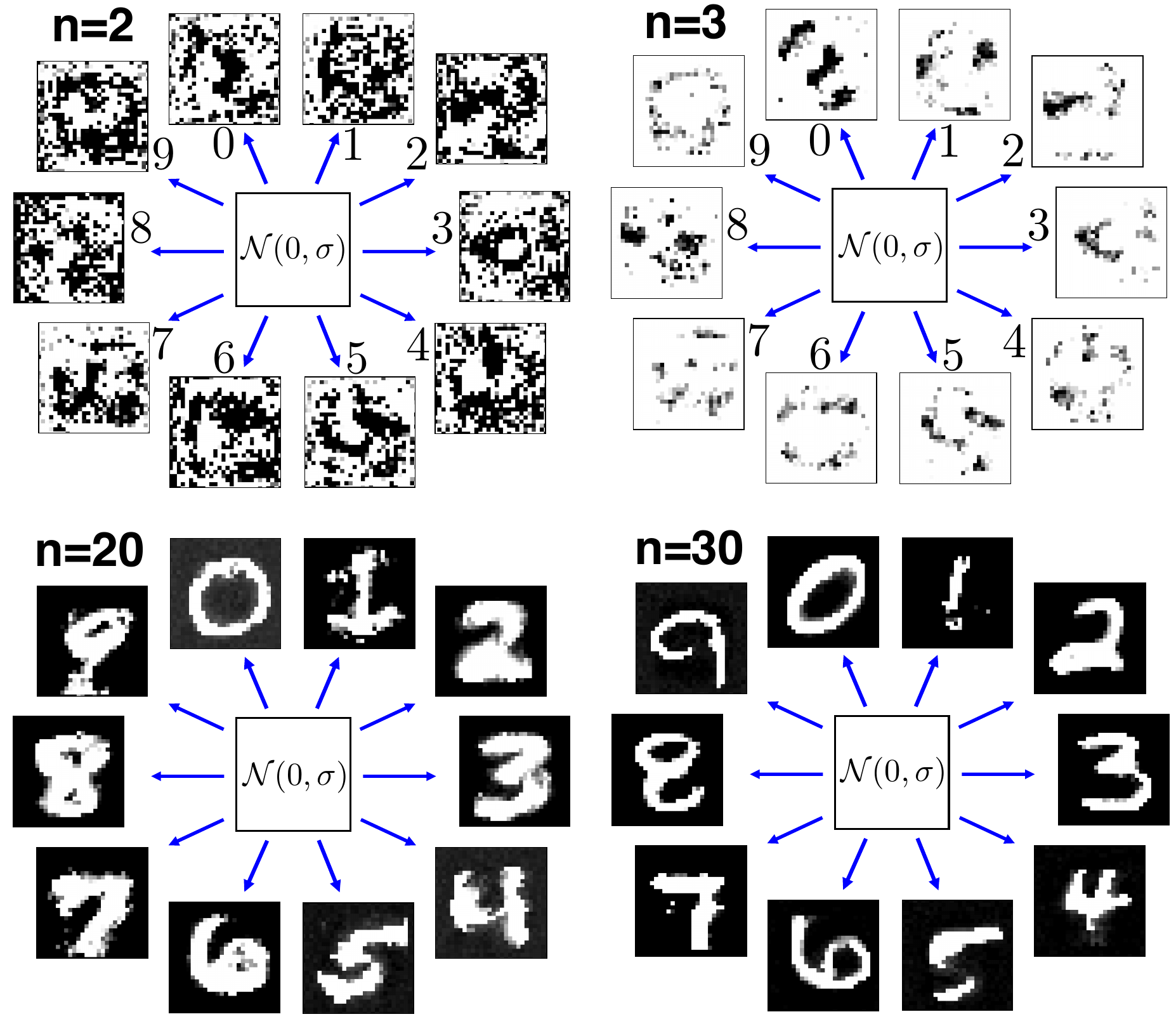}
\end{center}
\caption{\footnotesize{Examples of images generated from a gaussian noise ${\cal N} (0,0.1)$ by following the gradient of the 10 objective functions corresponding to 10 digit classes for model (\ref{update_rule}) with $n=2,3,20,30$. For $n=2$ and $n=3$ each final image is labeled. For $n=20$ and $n=30$ the labels are clear from the images. Throughout this paper we use grayscale intensities for representation of the image plane. They should not be confused with the color images, like the one in Fig.\ref{feature_to_prototype}, which shows the network's weights.}}\label{irreg_images}
\end{figure}

Once the training is complete, the neural network can be used to inspect the minima of the objective function. In order to do that one can define a set of $N_c=10$ objective functions each penalizing the deviations from the corresponding class
\begin{equation}
C^\alpha=\sum\limits_{\gamma=1}^{N_c} \Big(Y_\gamma - t_\gamma^{(\alpha)}\Big)^{2m}
\end{equation}
with the target output 
$$
t_\gamma^{(\alpha)}=\begin{cases} -1, \ \ \text{for}\ \  \gamma \neq \alpha\\
+1,  \ \ \text{for}\ \  \gamma=\alpha
\end{cases}
$$
A random image generated from a gaussian distribution can then be deformed into 10 images (sufficiently close to the initial one in the pixel space) corresponding to the 10 label classes by following the (negative) gradient of the objective functions according to the iterative rule 
\begin{equation}\label{gradient rule}
v_i \rightarrow v_i - \varepsilon u\Big( \frac{\partial C^\alpha}{\partial v_i}\Big),
\end{equation}
where $u\Big( \frac{\partial C^\alpha}{\partial v_i}\Big)$ is a unit vector, which points in the direction of the gradient of the objective function $C^\alpha$, and $\varepsilon$ is the size of the update step. This dynamics flows towards the minimum of the objective function, thus for a good learning algorithm we expect to see a recognizable image of the corresponding digit once it reaches the fixed point. The actual results are shown in Fig.\ref{irreg_images}. 
For small $n$ the dynamics converges to local minima which are not recognizable as digits by humans. These are the rubbish minima of \cite{Nguyen}. The training algorithm has learned them while trying to minimize the classification error on the training set. In contrast, for large $n$ the dynamics flows towards a minimum of the objective function that corresponds to a recognizable image of the corresponding digit. Thus this simple experiment demonstrates that for DAM with large $n$ the minima of the objective function have semantic meaning in the image space, while for models with small $n$ these minima are semantically meaningless. All this is achieved by training the network in a purely supervised way on the pixel permutation invariant task.

\section{Adversarial deformations}\label{adversarial deformation section}
A clean image, which is classified by a neural network as belonging to a certain class, can be modified by a small perturbation so that the deformed image is classified as a different class. What makes this statement a problem is that the perturbation, which is sufficient for changing the label, is small and typically is a speckled pattern that lacks semantic meaning. The most common technique for generating adversarial images is the sign of the gradient method of \cite{sign gradient}. For our purposes it is necessary to generate images that are placed exactly on the decision boundary between the classes and not just in its vicinity. For this reason we use a slightly different method. 

For each input image the vector of labels can be calculated by using (\ref{update_rule}). The elements of this vector can be sorted from largest to smallest value. The largest value of the output corresponds to the top classification choice of the network, the second largest output corresponds to the second choice, etc. 
\begin{figure}[h]
\begin{center}
\includegraphics[width = \linewidth]{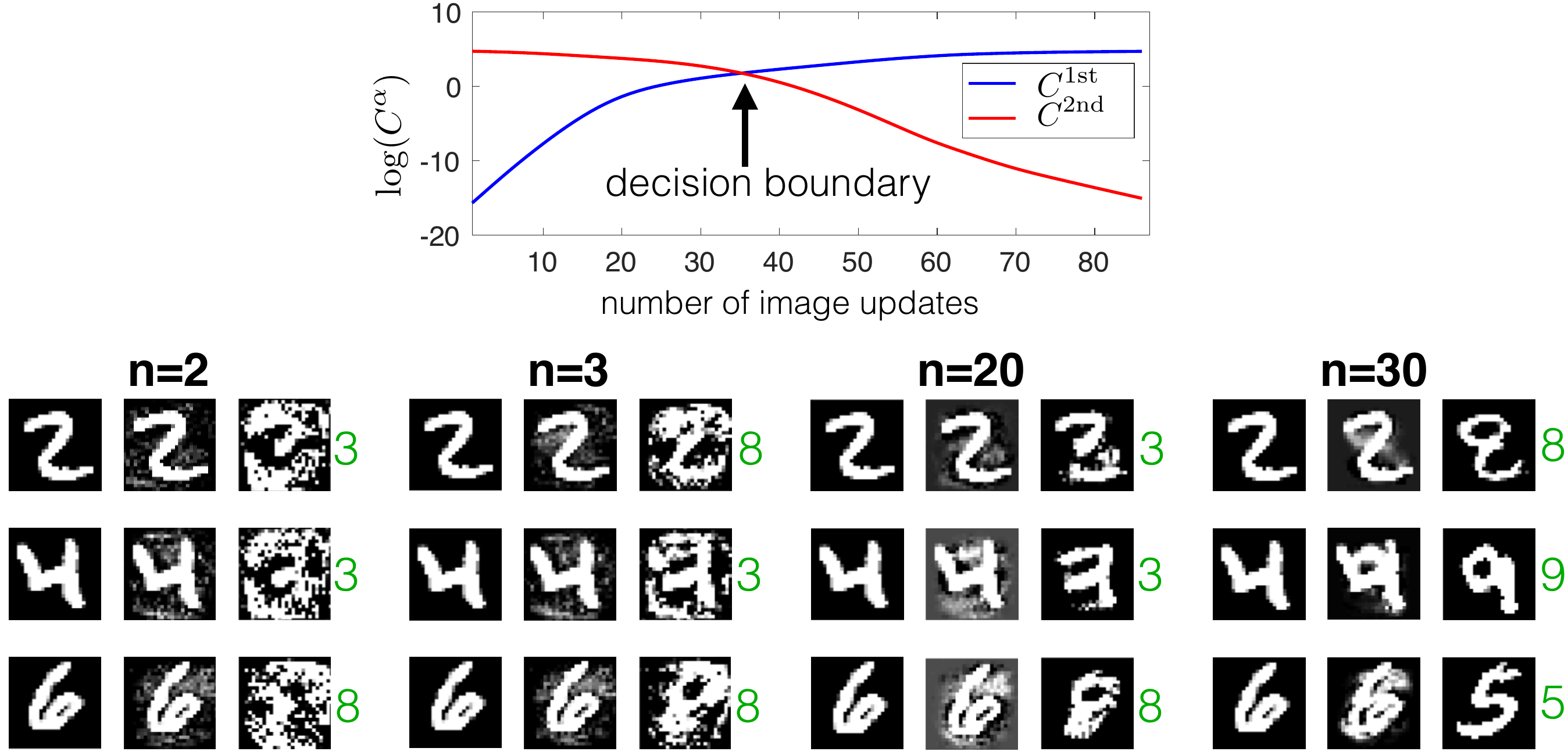}
\end{center}
\caption{\footnotesize{(Top panel) Logarithm of the top-choice and second-choice objective functions as the iterative dynamics following the negative gradient of the second-choice objective function (red) progresses. The crossing point defines the decision boundary. (Bottom panel) A set of triplets of images for $n=2, 3, 20, 30$, three per model. The first image in each triplet is a natural image from the dataset, the middle image is an artificial image corresponding to the crossing point of the two objective functions (decision boundary), the third image corresponds to the final point of the iterative dynamics, when the second-choice objective function (red curve) reaches zero.}}\label{adv_deform}
\end{figure}
The initial image can then be iteratively deformed along the (negative) gradient of the objective function corresponding to the second choice of the label made by the network. During this iterative process the first choice objective function will increase, and the second choice objective function will decrease (see top panel in Fig.\ref{adv_deform}). The image itself should gradually change from the initial clean image to an image that the network thinks should correspond to the second choice label. At some iteration the two objective functions become equal. This is the mathematical definition of the decision boundary. From the point of view of the neural network the image that corresponds to this point on the deformation trajectory is exactly in the middle between the two classes. The question is whether a human observer would agree with this interpretation, in other words whether or not this image looks ambiguous to humans.

 This method of generating the adversarial images has an advantage compared to \cite{sign gradient} since it guarantees that the image at the crossing point of the two objective functions is placed exactly on the decision boundary, and not just close to it. Also, in contrast to \cite{sign gradient}, it does not require a careful choice of the iteration step $\varepsilon$ in (\ref{gradient rule}), provided that it is sufficiently small. The procedure will simply need more steps in order to find the correct image corresponding to the crossing point, if the step is too small. The drawback of this method compared to \cite{sign gradient} is that it is slower.

In Fig.\ref{adv_deform} one can find a set of triplets of images generated by models with different values of power $n$. In each triplet, the first image is the initial clean image from the dataset, the second image is the one that the neural network has generated on the decision boundary, and the third one is the image that the network has generated when the iterative procedure (\ref{gradient rule}) has converged to the second choice label of the initial clean image. Two observations can be made from this figure. First, as the power of the interaction vertex is increased, the third image in each triplet becomes more semantically meaningful. This is in accord with the results of Fig.\ref{irreg_images}. Second, as the power of the interaction vertex is increased, the adversarial image (the middle image) on the decision boundary becomes more meaningful as well. For $n=2$ or $n=3$, the middle image looks almost the same as the initial clean image with a little bit of added speckled noise. The noise does not share much similarity with the second choice label. In contrast, for $n=30$, the deformation does not look like noise at all. The middle image looks as if the two digits (the initial seed and a digit from the target class) were morphed together in one image, or as if the second digit was added as a ``ghost'' image to the initial digit. These results suggest that models with large powers $n$ not only learn semantically meaningful set of minima of the objective function, but also learn semantically meaningful deformations between the classes. This is achieved by training the network on the pixel permutation invariant classification task, not by training it as a generative model. In principle, the target class does not have to necessarily coincide with the second choice of the initial classification decision. This convention is used for convenience. To the best of our knowledge, the results do not change qualitatively if instead of the second choice label, the initial image is deformed towards any other label but the top one.

\vspace{0.8cm}

\section{Transferability between the models with various $n$}
The most intriguing finding about rubbish and adversarial images is that they can be transferred between a diverse set of machine learning models. In order to test this phenomenon within the DAM framework we designed two experiments: one concerns the transfer of adversarial images and the other one concerns the transfer of rubbish images. 
\begin{figure}[t]
\begin{center}
\includegraphics[width = 0.7\linewidth]{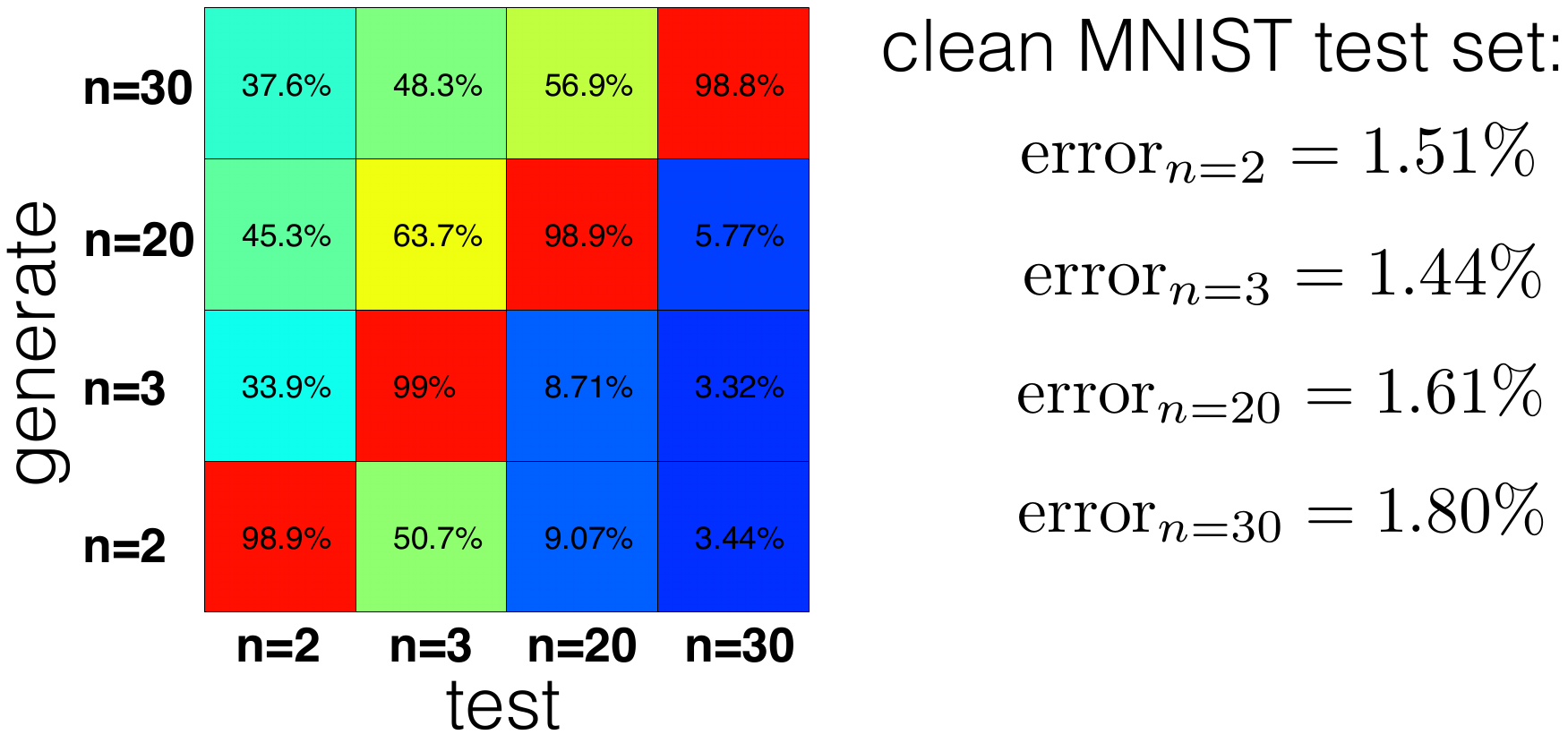}
\end{center}
\caption{\footnotesize{Transfer table of adversarial examples. The MNIST test set was used to construct four test sets of adversarial images, poised one step behind the decision boundary, for $n=2, 3, 20, 30$. These four datasets were  cross-classified by the same four models. Each dataset has 10000 images. The number at the intersection of the $i$-th row and the $j$-th column is the error rate of the $j$-th model on the adversarial dataset constructed using the $i$-th model. On the right the error rates of the four models on the original MNIST test set.}}\label{true_adv_transfer}
\end{figure}

In the first experiment the MNIST test set was used to generate adversarial examples by the four models with $n=2, 3, 20, 30$, resulting in four datasets each having 10000 images. For each clean image from the test set the procedure described in section \ref{adversarial deformation section} was used to generate an artificial image placed one iteration step behind the decision boundary defined in Fig.\ref{adv_deform}. The resulting image is thus classified as the second choice of the network on the original image. These four adversarial datasets were used for the cross-classification by the same four models. The error rates are shown in Fig.\ref{true_adv_transfer}. The diagonal elements of this table are close to $100\%$, which is guaranteed by the design of the datasets\footnote{The reason why the error rate is around 99\% instead of 100\% is because the classification error on the clean dataset is about 1.5\%, and in about 1\% of the cases the second choice of the network is the correct answer. Thus in these rare cases the adversarial deformation used to generate the dataset actually turns the incorrect answer into the correct one.}.  The most important aspect of this table is that the adversarial images generated by models with $n=2$ and $n=3$ do not transfer to the model with $n=30$. These images share semantic similarity with the clean image that was used as an initial seed for crafting them, but do not generally have semantic features of the target class of the deformation. The model with $n=30$ can detect this similarity with the initial image and can still correctly classify $97\%$ of these cases. In contrast the adversarial images crafted using the model with $n=30$ can be transferred to models with $n=2, 3$. However, as we argued in the previous section this is expected since these images share semantic similarities with both the initial seed and the target class of the deformation. Thus, any machine learning algorithm or a human subject should misclassify a substantial fraction of them.

\begin{figure}[h]
\begin{center}
\includegraphics[width = 0.99\linewidth]{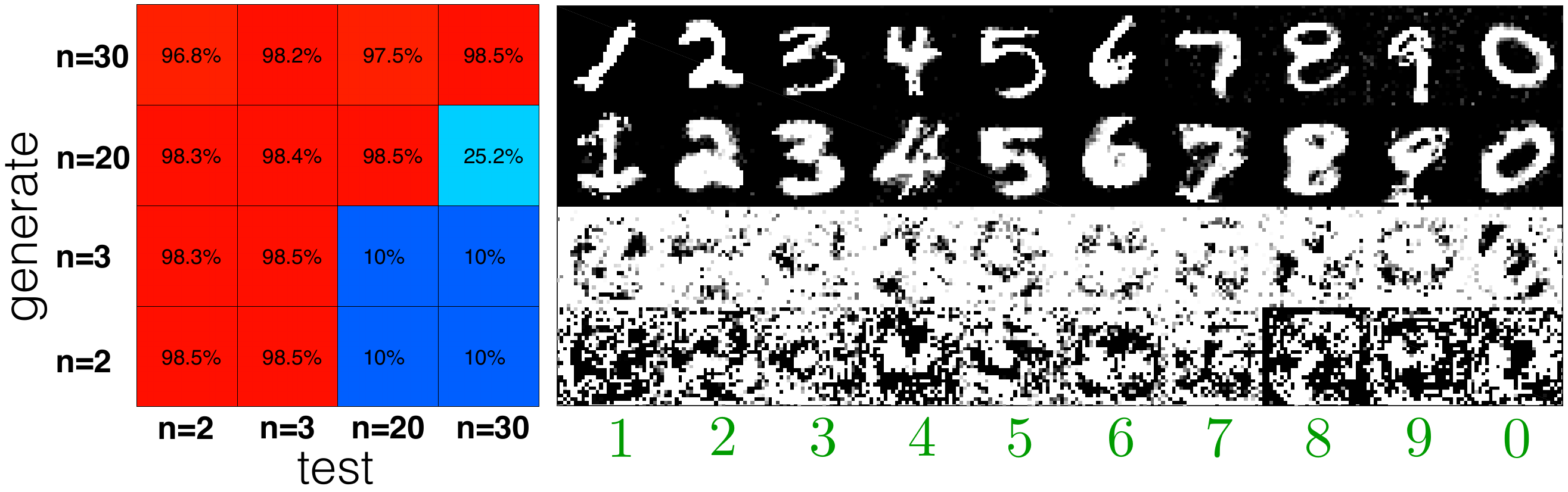}
\end{center}
\caption{\footnotesize{Transfer table. Each raw corresponds to a model that was used to generate an artificial data set (100 images per digit class, 1000 images in total) by following the gradient of the objective function. Each column corresponds to a model that was used to recognize these images. The number at the intersection is the mean confidence (the probability of the most probable label) produced by the test model. On the right one sample from the artificial dataset per label class. The labels of the images are shown in green at the bottom of each column. A single value $\beta_{\text{sm}}=3.2$ is used in this experiment. The results remain qualitatively similar if each model uses its own value of $\beta_{\text{sm}}$ in test time.}}\label{transfer_table}
\end{figure}
In the second experiment an artificial dataset was created, so that data correspond to the minima of the 10  objective functions $C^\alpha$. For each sample a random noise image was generated from a gaussian distribution, which was then iteratively changed in the direction of the negative gradient according to (\ref{gradient rule}) until convergence. The dataset has 100 images of each label class, 1000 images in total. This procedure was repeated for each value of $n=2, 3, 20, 30$. The resulting four datasets were used for cross-classification by the same four models. The results are shown in Fig.\ref{transfer_table}, but to discuss them we need to first introduce the notion of confidence. 

\subsection*{Confidence}
The output of the network (\ref{update_rule}) is a collection of $N_c$ numbers $-1\leq Y_\alpha\leq 1$. For the following discussion it is convenient to define a measure of confidence that the neural network has in making a classification decision. If the actual outputs $Y_\alpha$ are approximately equal to the target output, the confidence is high. In contrast, if the outputs $Y_\alpha$ are all approximately zero, with one being slightly larger than the rest $N_c-1$, the confidence is low. To quantify this effect, it is useful to pass the outputs through a softmax function to define a probability of each label given the input
$$
P_\alpha = \frac{\exp\big(\beta_{\text{sm}} Y_\alpha\big)}{\sum\limits_{\gamma=1}^{N_c} \exp\big(\beta_{\text{sm}} Y_\gamma\big)},
$$ 
where the parameter $\beta_{\text{sm}}$ is a number that regulates the steepness of the soft-max function. For each input image the confidence $Z$ of the network is defined as the probability of the most probable label. Thus if the network is confident that the presented image belongs to a certain class $Z \approx 100\%$, while if the network is unsure about what is shown in the image the distribution of labels is flat, and $Z \approx 10\%$ (in case of 10 classes for MNIST). The parameter $\beta_{\text{sm}}$ should be chosen in such a way that the average confidence of the network on a dataset matches its error on the test set. If the outputs of the network were exactly equal to the target outputs for all inputs, then the average confidence on a dataset would be equal to (since there is one correct choice and nine incorrect ones)
$$
Z=\frac{e^{\beta_{\text{sm}}}}{e^{\beta_{\text{sm}}} + 9 e^{-\beta_{\text{sm}}}} \approx 0.985
$$
The error rate of all our models is of the order of $1.5\%$, which defines the right hand side of the above equation. This results in $\beta_{\text{sm}}\approx 3.2$.  Since the outputs of the networks are in general slightly different from the target outputs, the actual values of $\beta_{\text{sm}}$ are different for the four models that are discussed. They are: $\beta_{\text{sm}}^{n=2}=4.7$, $\beta_{\text{sm}}^{n=3}=4.6$, $\beta_{\text{sm}}^{n=20}=4$, $\beta_{\text{sm}}^{n=30}=26$. However, all the conclusions presented below remain true, for any values of $\beta_{\text{sm}}$ as long as they stay within the range $2\leq\beta_{\text{sm}}\leq 10^7$.  

\begin{figure}[h]
\begin{center}
\includegraphics[width = 0.5\linewidth]{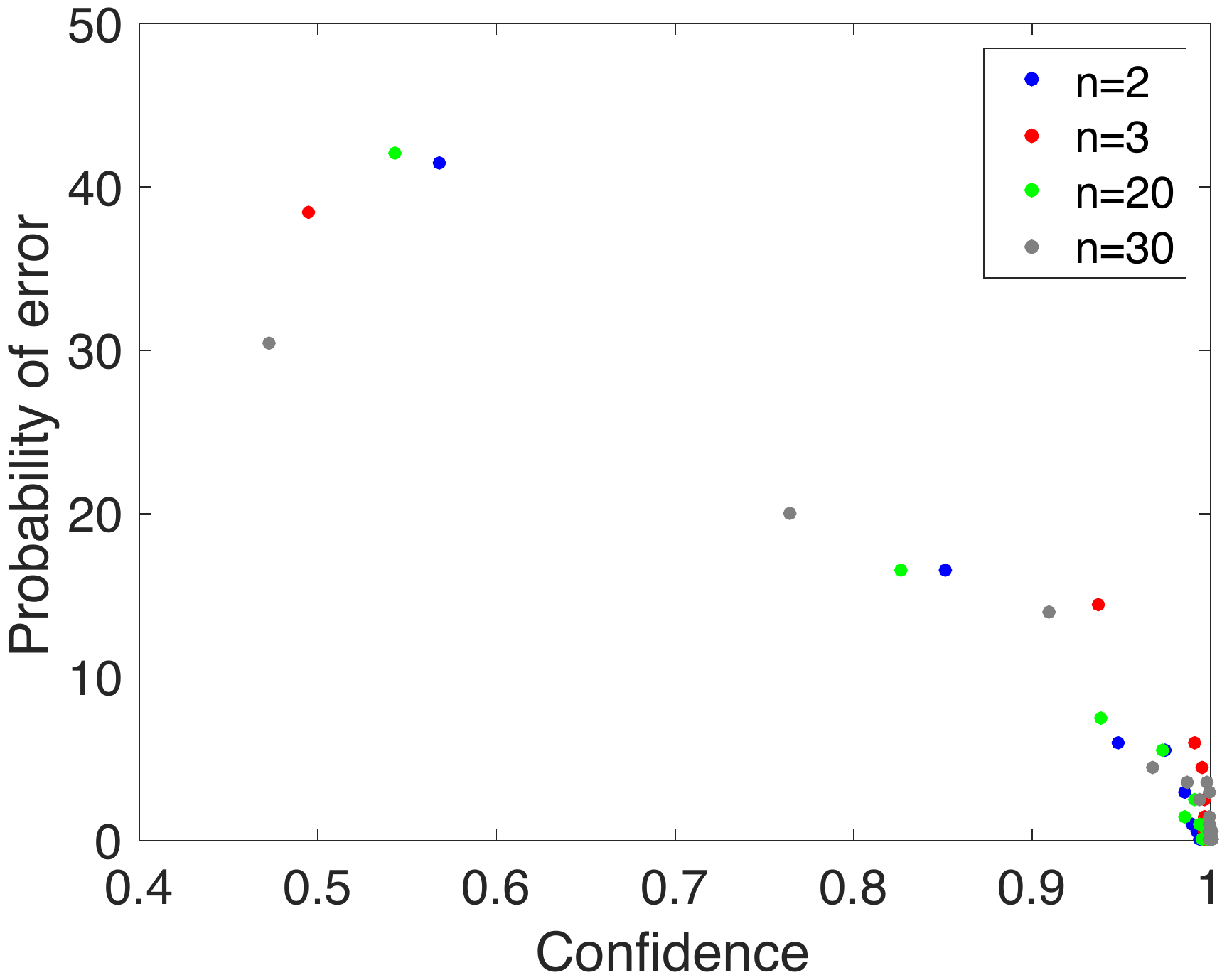}
\end{center}
\caption{\footnotesize{The 10000 images from the MNIST test set are binned in groups having a certain value of confidence. The classification error for each group is calculated. The values of the slope are: $\beta_{\text{sm}}^{n=2}=4.7$, $\beta_{\text{sm}}^{n=3}=4.6$, $\beta_{\text{sm}}^{n=20}=4$, $\beta_{\text{sm}}^{n=30}=26$.}}\label{error_vs_confidence}
\end{figure}
The images from the MNIST test set can be binned into groups having certain confidence and the classification error can be calculated on each bin. The results are shown in Fig.\ref{error_vs_confidence}. The probability of error decreases as the confidence increases, a result suggesting that the confidence measure is a meaningful quantity. Now we return to the discussion of the second experiment pertaining to the analysis of ``rubbish'' images.

In Fig.\ref{transfer_table} the mean confidence of the 16 cross-classification pairs is reported together with 10 sample images from each dataset, one for each class, for the four data sets. The bottom row corresponds to the data generated by the model with $n=2$. The samples look like rubbish images, in accord with Fig.\ref{irreg_images}. These rubbish images are recognized by the network with $n=2$ at the average confidence $98.5\%$, which is the result of the particular choice of the value $\beta_{\text{sm}}$. These rubbish images can be transferred to the model with $n=3$, which is also very confident that they correspond to the correct labels. However, if one tries to transfer these rubbish images to the models with $n=20$ or $n=30$, these higher order models immediately detect that these images do not correspond to any class and produce an average confidence $\approx 10\%$. This accurately mimics the behavior of a human subject who would immediately detect that these rubbish images are semantically meaningless, thus should not belong to any class.  The same conclusion applies to the dataset constructed by the model with $n=3$. In contrast, the dataset constructed by the model with $n=30$ is composed of nice images of digits, which are recognizable by humans. These images can be transferred to any model with $n\leq 30$ without loss in confidence. These results suggest that models with large $n$ better mimic human visual cognition, compared to models with small $n$, both at generation and at testing. 

From the security perspective these results suggest that the models with large $n$ can be used to detect and stop a potential hacker attack, which is devised by exploiting conventional machine learning techniques (for example DNNs with ReLUs). At the same time, if the adversary tries to use models with large $n$ for mounting the attack and deploys it against models with any (small or large) $n$, this does not possess a security issue, because the interpretations that any model gives to these images are consistent with human visual perception. 

From this perspective the models with large $n$ can be valuable even if their classification accuracy is slightly lower than that of models with small $n$. They can be used in pair with another model that has a good classification accuracy, but that is vulnerable to adversarial/rubbish examples. The first model (with large $n$) detects a potential adversarial attack. In cases when it is confident, and its label disagrees with the other more accurate model, one can use the prediction of the more accurate model for the final classification decision. However if the large $n$ model is unconfident of a particular input, that input should be labeled as junk and not processed by the more accurate/vulnerable model.

\section{Discussion and conclusions}
Although modern machine learning techniques outperform humans on many classification tasks, there is a serious concern that they do not understand the structure of the training data. A clear demonstration of this lack of understanding was presented in \cite{Szegedy,Nguyen}, who showed two examples of nonsensical predictions of DNNs that contradict to human visual perception: adversarial images and rubbish images. In the present paper we propose that DAM with higher order interactions in the energy function produce more sensible interpretations (consistent with human vision) of adversarial and rubbish images. We argue that these models better mimic human visual perception than DNNs with ReLUs. 

A possible explanation of adversarial examples, pertaining to neural networks being too linear, was given in \cite{sign gradient}. Our explanation follows the same line of thought with some differences. One result of \cite{Krotov Hopfield} is that DAMs with large powers of the interaction vertex in the energy function are dual to feed-forward neural nets with highly non-linear activation functions - the rectified polynomials of higher degrees. From the perspective of this duality, one might expect that by simply replacing the ReLUs in DNNs by the higher rectified polynomials one might solve the problem of adversarial and rubbish images for a sufficiently large power of the activation function. We tried that and discovered that although DNNs with higher rectified polynomials alone perform better than DNNs with ReLUs from the adversarial perspective, they are worse than DAMs with the update rule (\ref{update_rule}). These observations need further comprehensive investigation. Thus, simply changing ReLUs to higher rectified polynomials is not enough to get rid of adversarial problems, and other aspects of the training algorithm presented in section \ref{DAM networks} are important. 

For thinking about neurobiology the energy functions with higher order interactions considered in \cite{Krotov Hopfield} and in this paper should be thought of as effective theories that arise after excluding the auxiliary variables from the microscopic description. The microscopic realization in terms of biological neurons will be discussed elsewhere. 

There are two straightforward ideas for possible extensions of this work. First, it would be interesting to complement the proposed training procedure with adversarial training. In other words to train the network using the algorithm of section \ref{DAM networks} but on a combination of clean images and adversarial images, along the lines of \cite{Nokland,sign gradient}.  We expect that this should further increase the robustness to adversarial examples and increase the classification accuracy on the clean images. Second, it would be interesting to investigate the proposed methods in the convolutional setting. Naively, one expects that the adversarial problems are more severe in the fully connected networks than in the convolutional networks. For this reason we used the fully connected networks for our experiments. We expect that the training algorithm of section \ref{DAM networks} can be combined with convolutional layers to better describe images.  

Although more work is required to fully resolve the problem of adversarial and rubbish images, we believe that the present paper has identified a promising computational regime of the neural networks that significantly mitigates the vulnerability of DNNs to adversarial and rubbish images and that remains little investigated.

%----------------------------------------------------------------------------------------

\end{document}